\documentclass[10pt,twocolumn,letterpaper]{article}
\usepackage[pagenumbers]{cvpr}
\usepackage{amsmath,amssymb,amsfonts}
\usepackage{algorithmic}
\usepackage{graphicx}
\usepackage{textcomp}
\usepackage{xcolor}
\usepackage{url}
\usepackage{algorithm}
\usepackage{algorithmic}
\usepackage{amsfonts}
\usepackage{amsmath}
\usepackage{marvosym}
\usepackage{booktabs}
\usepackage{pifont}
\usepackage{multirow}
\newcommand{\settablefont}{\fontsize{8}{15}\selectfont}
\usepackage{makecell}

\usepackage{newfloat}
\usepackage{listings}
\usepackage{color}

\definecolor{pa}{rgb}{0.0, 0.7529411764705882, 0.5568627450980392}
\definecolor{dp}{rgb}{0.3568627450980392, 0.607843137254902, 0.8352941176470588}
\definecolor{ta}{rgb}{0.8784313725490196, 0.3882352941176471, 0.4666666666666667}
\definecolor{sg}{rgb}{0.5411764705882353, 0.1686274509803922, 0.8862745098039216}

\definecolor{cvprblue}{rgb}{0.21,0.49,0.74}
\usepackage[pagebackref,breaklinks,colorlinks,allcolors=cvprblue]{hyperref}

\title{ViPOcc: Leveraging Visual Priors from Vision Foundation Models for Single-View 3D Occupancy Prediction}

\author{
	Yi Feng$^{1}$,\and Yu Han$^2$,\and   Xijing Zhang$^1$,\and   Tanghui Li$^1$,\and   Yanting Zhang$^2$,\and Rui Fan$^{1}$\thanks{Corresponding author: Rui Fan.}
        \and
	$^1$Tongji University,
        \\
        $^2$Donghua University
        \\
        {\tt \{fengyi, rui.fan\}@ieee.org}
}
\begin{document}

\maketitle

\begin{abstract}
Inferring the 3D structure of a scene from a single image is an ill-posed and challenging problem in the field of vision-centric autonomous driving. Existing methods usually employ neural radiance fields to produce voxelized 3D occupancy, lacking instance-level semantic reasoning and temporal photometric consistency. In this paper, we propose ViPOcc, which leverages the visual priors from vision foundation models (VFMs) for fine-grained 3D occupancy prediction. Unlike previous works that solely employ volume rendering for RGB and depth image reconstruction, we introduce a metric depth estimation branch, in which an inverse depth alignment module is proposed to bridge the domain gap in depth distribution between VFM predictions and the ground truth. The recovered metric depth is then utilized in temporal photometric alignment and spatial geometric alignment to ensure accurate and consistent 3D occupancy prediction. Additionally, we also propose a semantic-guided non-overlapping Gaussian mixture sampler for efficient, instance-aware ray sampling, which addresses the redundant and imbalanced sampling issue that still exists in previous state-of-the-art methods. Extensive experiments demonstrate the superior performance of ViPOcc in both 3D occupancy prediction and depth estimation tasks on the KITTI-360 and KITTI Raw datasets. Our code is available at: \url{https://mias.group/ViPOcc}.
\end{abstract}

\section{Introduction}
\label{sec.intro}
As a key ingredient of environmental perception in autonomous driving, 3D occupancy prediction has garnered considerable attention in recent years \cite{cao2022monoscene, wei2023surroundocc, huang2024selfocc, tian2024occ3d}. Early efforts tackle this problem through supervised learning, which requires extensive 3D human-labeled annotations and depth ground truth acquired using additional range sensors \cite{huang2023tri}. More recently, neural radiance field (NeRF)-based approaches have emerged as promising techniques for unsupervised single-view 3D occupancy prediction \cite{behind2023wimbauer, know2024li}, noted for their capability to render photorealistic images from novel viewpoints. 
\begin{figure}[!t]
	\centering
	\includegraphics[width=0.9999\linewidth]{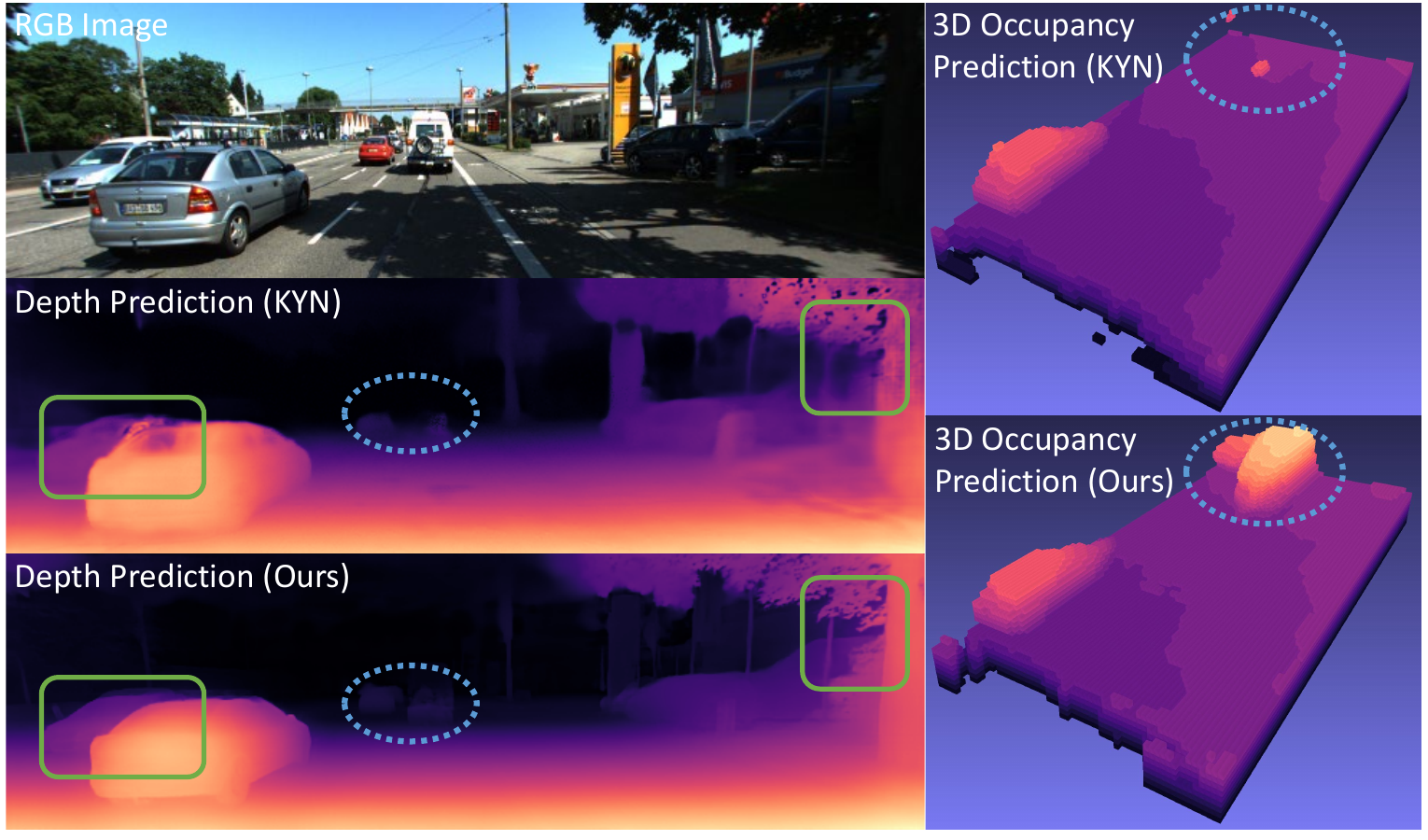}
	\caption{\textbf{Single-view 3D scene reconstruction results.} KYN \cite{know2024li} struggles to recover clear object boundaries (green boxes) and exhibits poor reconstruction performance for distant objects (blue circles). ViPOcc outperforms KYN in both monocular depth estimation and 3D occupancy prediction tasks.}
	\label{fig.teaser}
\end{figure}

As a pioneering work, BTS \cite{behind2023wimbauer} estimates a 3D density field from a single view, relying solely on photometric consistency constraints across multiple views during training. Subsequent studies \cite{han2024boosting, know2024li} adopt the same training strategy for 3D scene reconstruction but often underexploit temporal photometric and geometric constraints, resulting in inconsistent 3D occupancy predictions across adjacent frames.

Another growing trend is to unleash the potential of vision foundation models (VFMs) for comprehensive 3D scene representation. As a notable example, KYN \cite{know2024li} leverages a large vision-language model to enrich 3D features with semantic information. However, as illustrated in Fig. \ref{fig.teaser}, challenges remain, particularly with the frequent omission of critical instances, due to the indiscriminate random ray sampling process. SC-DepthV3 \cite{sun2023sc} uses predictions from LeReS \cite{yin2021learning} as pseudo depth for robust unsupervised depth estimation. Nevertheless, current VFMs generally produce monocular depth predictions with inherent scale ambiguity \cite{yang2024depth}, which are not directly applicable to temporal photometric alignment. The recently proposed VFM, Depth Anything V2 \cite{yang2024depthv2}, demonstrates exceptional zero-shot performance in metric depth estimation with fine-grained details. However, it experiences a significant performance decline due to domain discrepancies between the training and test data.

To address the aforementioned challenges, we introduce ViPOcc, a novel approach that leverages visual priors from VFMs for fine-grained, instance-aware 3D scene reconstruction. Unlike previous state-of-the-art (SoTA) methods that solely utilize photometric discrepancies as supervisory signals, our method incorporates a depth prediction branch, which fully exploits inter-frame photometric consistency and intra-frame geometric reconstruction consistency, enabling self-supervised training with spatial-temporal consistency constraints. 

Specifically, we design an inverse depth alignment module that mitigates the discrepancies between VFM predictions and depth ground truth, leading to compelling metric depth estimation results. To further enhance both the efficiency and accuracy of 3D occupancy prediction, we develop a semantic-guided, non-overlapping Gaussian mixture (SNOG) sampler, which effectively addresses issues such as redundant ray sampling and the overlooking of crucial instances prevalent in previous methods. Additionally, we propose a temporal alignment loss and a reconstruction consistency loss, which further improve the quality of both metric depth and 3D occupancy predictions. Extensive experiments on the KITTI-360 and KITTI Raw datasets validate the effectiveness of each developed component and further demonstrate ViPOcc's superior performance over all existing SoTA methods.

In a nutshell, we present the following key contributions: 
\begin{enumerate}
\item We propose \textbf{ViPOcc}, a single-view 3D \textbf{Occ}upancy prediction framework that incorporates \textbf{Vi}sual \textbf{P}riors from VFMs, achieving SoTA performance in both monocular depth estimation and 3d occupancy prediction tasks.

\item We introduce an inverse depth alignment module that effectively recovers the scale of the VFM's depth predictions while preserving their local visual details.

\item We present a SNOG sampler that guides the framework to focus more on crucial instances and avoid overlapping patches during ray sampling.

\item We establish a novel training paradigm that couples the unsupervised training of 3D occupancy prediction and monocular depth estimation using the proposed temporal alignment and reconstruction consistency losses.

\end{enumerate}

\begin{figure*}[!t]
	\centering
	\includegraphics[width=0.99\linewidth]{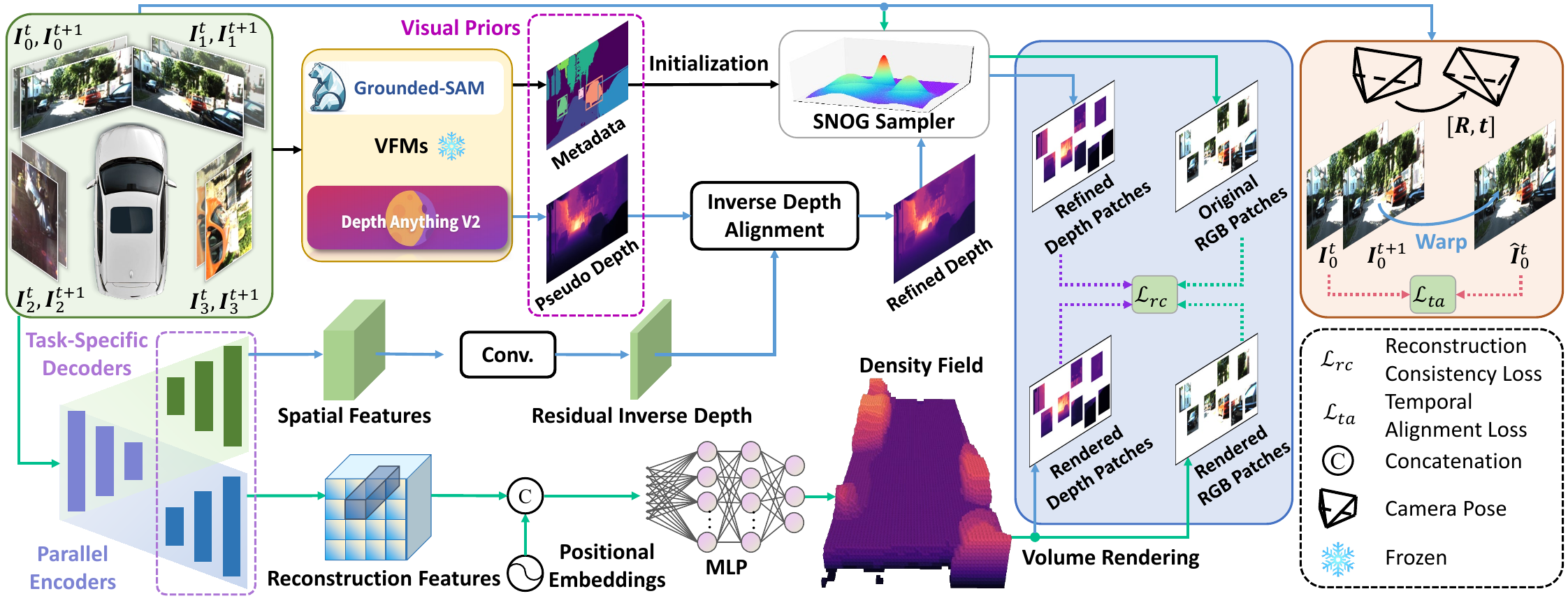}
	\caption{An illustration of our proposed \textbf{ViPOcc} framework. Unlike \textcolor{pa}{{previous approaches}} that rely solely on NeRF for 3D scene reconstruction, ViPOcc introduces an additional \textcolor{dp}{{depth prediction branch}} and an instance-aware SNOG sampler for \textcolor{ta}{{temporal photometric alignment}} and \textcolor{sg}{{spatial geometric alignment}}.}
	\label{fig.overview}
\end{figure*}
\section{Related Work}
\subsection{Single-View 3D Occupancy Prediction}
Deriving voxelized 3D occupancy of a scene from a single image is a promising technique for achieving fine-grained geometric representation and comprehensive environmental understanding in 3D space \cite{zhang2024vision}. As a pioneering work, MonoScene \cite{cao2022monoscene} leverages voxel features generated through view projection for occupancy regression. However, this method is not suitable for real-time multi-view 3D reconstruction due to the inefficiency of voxel representations. TPVFormer \cite{huang2023tri} extends it to a multi-camera setup by incorporating tri-perspective view representations. Despite their compelling performance, these supervised methods necessitate data with 3D ground truth, which requires labor-intensive human annotation. Recently, the study \cite{behind2023wimbauer} introduced BTS, a fully unsupervised method that uses perspective and fisheye video sequences to reconstruct driving scenes with NeRF-based volume rendering techniques. Following this work, KYN \cite{know2024li} leverages meaningful semantic and spatial context for fine-grained 3D scene reconstruction. MVBTS \cite{han2024boosting} combines density fields from multi-view images through knowledge distillation, achieving SoTA performance in handling occluded regions. Different from existing NeRF-based frameworks, we incorporate an additional depth prediction branch for spatial-temporal 3D occupancy alignment.

\subsection{Visual Priors for 3D Scene Reconstruction}
Previous studies \cite{know2024li, zhang2023occnerf} have integrated visual priors from pre-trained VFMs into depth estimation and NeRF-based 3D scene reconstruction frameworks. Existing depth estimation methods typically utilize pre-inferred semantics for fine-grained feature representation and fusion \cite{guizilini2020semantically, jung2021fine, chen2023self}. Other studies \cite{kerr2023lerf, peng2023openscene} leverage 2D visual priors for 3D feature representation and registration. KYN \cite{know2024li} incorporates a pre-trained vision-language network for robust 3D feature representation, significantly improving 3D shape recovery. MonoOcc \cite{zheng2024monoocc} employs a pre-trained InternImage-XL \cite{wang2023internimage} as its backbone for visual feature extraction and distillation. OccNeRF \cite{zhang2023occnerf} utilizes frozen VFMs for 2D semantic supervision but faces challenges in detecting small instances due to the limitations of open-vocabulary models in capturing fine details. While these methods have successfully leveraged the strengths of VFMs for feature extraction, the informative visual priors from VFMs remain underutilized. In this paper, we leverages semantic priors from Grounded-SAM \cite{ren2024grounded} and spatial priors from Depth Anything V2 \cite{yang2024depthv2} for efficient ray sampling and spatial-temporal 3D occupancy alignment.

\subsection{Unsupervised Monocular Depth Estimation}
Existing frameworks typically maximize photometric consistency across video sequences or stereo image pairs to estimate scale-invariant depth maps. SfMLearner \cite{zhou2017unsupervised}, the first reported study in this field, jointly estimates depth maps and camera poses between successive video frames by minimizing a photometric reprojection loss. Building on this method, Monodepth2 \cite{godard2019digging} introduces a minimum reprojection loss to address occlusion issues and an automasking loss to exclude moving objects that appear stationary relative to the camera. Subsequent studies mainly explored various network architectures \cite{wang2024sqldepth, watson2021temporal}, dynamic object filtering strategies \cite{sun2024dynamo, yin2018geonet}, and additional constraints \cite{guizilini2020semantically, schmied2023r3d3}. Other NeRF-based frameworks \cite{behind2023wimbauer, han2024boosting} estimate metric depth maps through discrete volume rendering. However, their predictions often lack accuracy and fail to preserve clear object contours. In contrast, our proposed ViPOcc utilizes visual priors from VFMs to enable instance-aware ray sampling and fine-grained metric depth estimation.

\section{Methodology}
\label{sec.method}
\subsection{Problem Setup}
Given an input RGB image $\boldsymbol{I}$ and its corresponding intrinsic matrix $\boldsymbol{K}$, we aim to reconstruct the 3D geometry of the entire scene with the voxelized density:
\begin{equation}
	 \sigma_{\boldsymbol{p}} = \mathcal{R}(\boldsymbol{p}, \boldsymbol{I}, \boldsymbol{K}, \boldsymbol{\Theta}),
\end{equation}
where $\boldsymbol{p}$ denotes a 3D point in the reconstructed scene, and $\mathcal{R}(\cdot)$ represents the neural radiance field with learnable parameters $\boldsymbol{\Theta}$. $\sigma_{\boldsymbol{p}}$ can be further employed to produce a rendered RGB image $\hat{\boldsymbol{I}}_r$ and a rendered distance map $\hat{\boldsymbol{D}}_r$ using the following expressions:
\begin{equation}
	\hat{\boldsymbol{I}}_r(\boldsymbol{p}_i) = \sum_{i=1}^{M}T_i\alpha_i c_{\boldsymbol{p}_i}, \ \ \ 
	\hat{\boldsymbol{D}}_r(\boldsymbol{p}_i) = \sum_{i=1}^{M}T_i\alpha_id_i, 
\end{equation}
where $\alpha_i = 1 - \exp{(-\sigma_{\boldsymbol{p}_i}||\boldsymbol{p}_{i+1}-\boldsymbol{p}_i||_2)}$ denotes the probability that the ray ends between $\boldsymbol{p}_i$ and $\boldsymbol{p}_{i+1}$, $T_i = \prod_{j=1}^{i-1}(1-\alpha_j)$ represents the accumulated transmittance, $c_{\boldsymbol{p}_i}$ denotes the sampled RGB value from other viewpoints, and $d_i$ represents the distance between $\boldsymbol{p}_i$ and the ray origin.

\subsection{Architecture Overview}
As illustrated in Fig. \ref{fig.overview}, ViPOcc takes stereo image pairs $\boldsymbol{I}_{0,1}^{t,t+1}$ and rectified fisheye images $\boldsymbol{I}_{2,3}^{t,t+1}$ captured at timestamps $t$ and $t+1$ as input. $\boldsymbol{I}_{0}^t$ is regarded as the principal frame, from which spatial features $\boldsymbol{F}_s$ and reconstruction features $\boldsymbol{F}_r$ are extracted using parallel encoders and task-specific decoders. During training, ViPOcc simultaneously generates 2D depth maps and 3D density fields from two separate branches. In the depth estimation branch, an inverse depth alignment module is designed to mitigate the domain discrepancy between depth priors from a VFM and the depth ground truth. The refined depth maps $\hat{\boldsymbol{D}}$ and the corresponding RGB images are then fed into our developed SNOG sampler for efficient ray sampling, producing instance-aware and non-overlapping patches. On the other hand, in the 3D occupancy prediction branch, $\boldsymbol{F}_r$ combined with positional embeddings $\boldsymbol{F}_{p}$ is passed through an MLP to predict a 3D density field, which is then utilized in volume rendering to generate depth and RGB patches. By enforcing reconstruction consistency across sampled RGB and depth patches, as well as temporal photometric consistency between adjacent principal frames, we achieve improved performance in both 3D occupancy prediction and metric depth estimation. 

\subsection{Inverse Depth Alignment}
\label{sec.method_ida}
Unlike prior arts \cite{behind2023wimbauer, han2024boosting} that rely solely on NeRF-based reconstruction consistency to supervise framework training, we incorporate inter-frame photometric consistency and depth rendering consistency through a VFM-driven depth estimation branch. Pseudo depth maps $\boldsymbol{D}_{p}$ are first obtained from off-the-shelf VFMs like Depth Anything V2 \cite{yang2024depthv2}. 
Nevertheless, as demonstrated in our experiments, the residuals between pseudo and ground-truth depth data exhibit dramatically deviated distributions. These deviations arise from significant domain gaps between real-world scenarios and the data on which VFMs are initially trained. Therefore, it is imperative to refine depth before utilizing it to introduce additional constraints for temporal photometric alignment. As discussed in \cite{he2016deep}, neural networks often struggle to converge or maintain accuracy when fitting large ranges of numerical variations. It is thus plausible to fit residual inverse depth in our task, expressed as $\mathcal{F}(\boldsymbol{x}):= \frac{1}{\hat{\boldsymbol{D}}(\boldsymbol{x})} - \frac{1}{\boldsymbol{D}_{p}(\boldsymbol{x})}$, where $\mathcal{F}(\cdot)$ denotes the residual inverse depth function, $\hat{\boldsymbol{D}}$ represents the refined depth map, and $\boldsymbol{x}$ denotes a given 2D pixel. This function can be effectively fitted using spatial features $\boldsymbol{F}_s$ by formulating it as $\mathcal{F}(\boldsymbol{x}) = f(\boldsymbol{F}_s, \boldsymbol{\theta})$, where $f(\cdot)$ denotes a convolutional layer with learnable parameters $\boldsymbol{\theta}$. The refined depth map can therefore be yielded as follows:
\begin{equation}
	\hat{\boldsymbol{D}}(\boldsymbol{x}) = \frac{1}{\frac{1}{\boldsymbol{D}_{p}(\boldsymbol{x})} + f(\boldsymbol{F}_s, \boldsymbol{\theta}) + \epsilon},
\end{equation}
where $\epsilon$ is a small constant used to prevent the denominator from being zero. $\hat{\boldsymbol{D}}$ can then be used to ensure inter-frame photometric consistency and depth rendering consistency.

\subsection{Semantic-Guided Non-Overlapping Gaussian Mixture Sampler}

\begin{figure}[!t]
	\centering
	\includegraphics[width=0.99\linewidth]{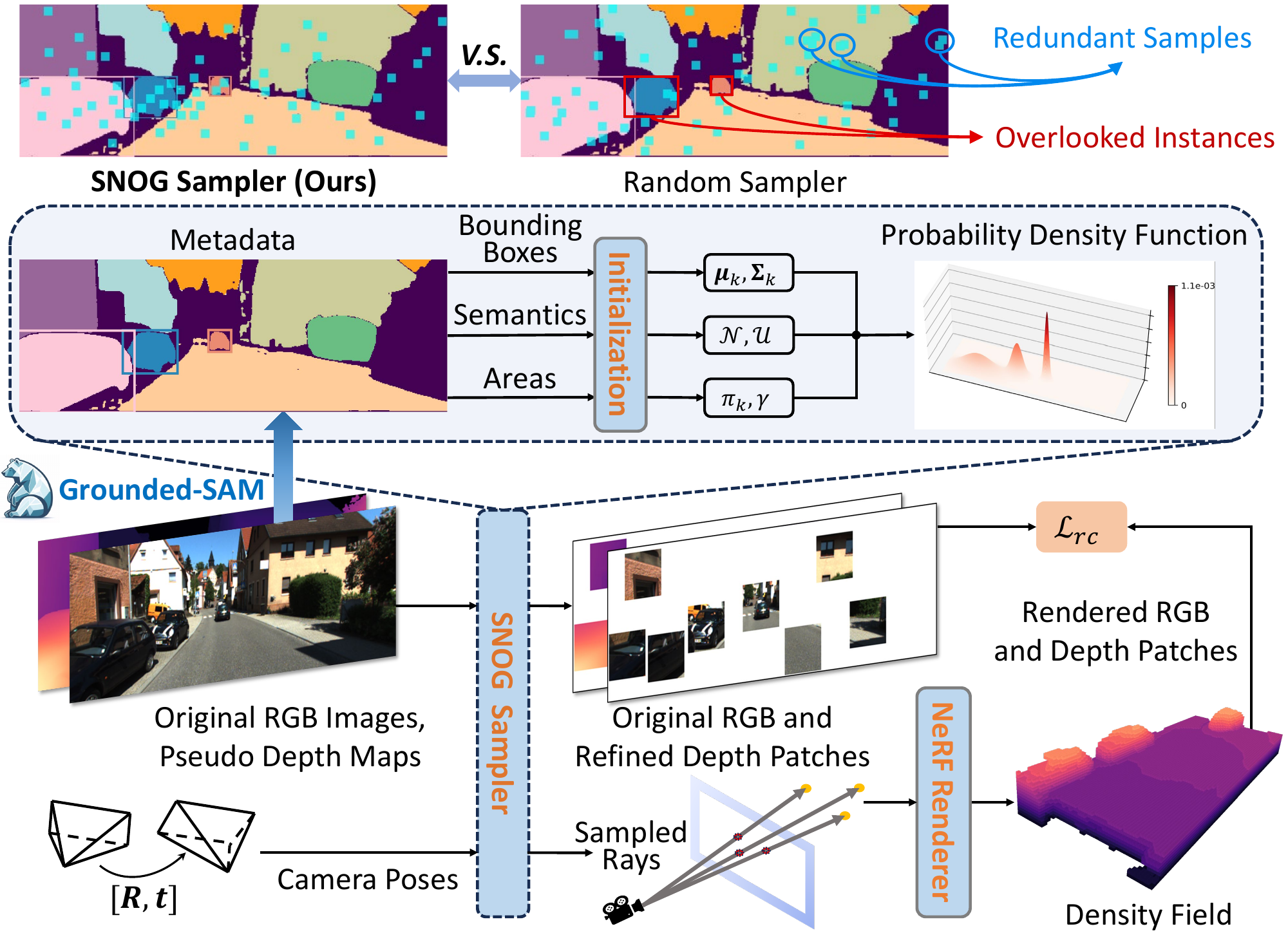}
	\caption{An illustration of our proposed SNOG sampler.}
	\label{fig.sampler}
\end{figure}

Focusing on individual instances rather than the entire scene can lead to more detailed and fine-grained 3D scene reconstruction. However, as shown in Fig. \ref{fig.sampler}, previous SoTA approaches \cite{behind2023wimbauer, know2024li} typically adopt a random patch sampler for uniform ray sampling across the entire scene, leading to redundant samples and overlooked instances. In contrast, our proposed SNOG sampler leverages informative visual priors from the pre-trained open-vocabulary model Grounded-SAM \cite{ren2024grounded} (a combination of Grounding DINO \cite{liu2023grounding} and SAM \cite{kirillov2023segment}) to optimize the allocation of computational resources while enhancing the awareness of crucial instances. 

Specifically, we utilize the semantic labels from the Cityscapes dataset \cite{cordts2016cityscapes} as prompts for Grounding DINO. After obtaining instance-level bounding boxes, we employ SAM to generate precise segmentation masks. Consequently, for the $k$-th instance, we acquire its metadata $\mathcal{M}_k=\{\boldsymbol{l}_k, \boldsymbol{b}_k, s_k\}$, where $\boldsymbol{l}_k$ denotes the center location of its bounding box, $\boldsymbol{b}_k$ stores half of the height and width of its bounding box, and $s_k$ indicates the semantic area of the instance. Subsequently, we use Gaussian mixture distribution combined with background uniform distribution to achieve instance-aware and non-overlapping ray sampling, the probability density function (PDF) $p(\boldsymbol{x})$ of which can be formulated as follows:
\begin{equation}
	p(\boldsymbol{x}) = 
	(1-\gamma)\sum_{k=1}^{K} \pi_{k} \mathcal{N}\left(\boldsymbol{x} \mid \boldsymbol{\mu}_{k}, \boldsymbol{\Sigma}_{k}\right) + 
	\gamma\mathcal{U}\left(\boldsymbol{x} \mid s\right), 
 \label{eq.pdf}
\end{equation}
where $\mathcal{N}\left(\boldsymbol{x} \mid \boldsymbol{\mu}_{k},\boldsymbol{\Sigma}_{k}\right)$ denotes the PDF of the bivariate normal distribution with mean vector $\boldsymbol{\mu}_{k}$ and covariance matrix $\boldsymbol{\Sigma}_{k}$, $\mathcal{U}(\boldsymbol{x}\mid s)$ denotes the PDF of a 2D uniform distribution within the area $s$, and $\gamma$ and $\pi_k$ denote the weights of the background sampling and independent Gaussian distributions, respectively. 

For the Gaussian distribution of the $k$-th instance, our objectives are to 1) locate $\boldsymbol{\mu}_k$ at the center of its bounding box and 2) ensure that approximately 95.5\% of the samples fall within the bounding box. We can therefore initialize the parameters in (\ref{eq.pdf}) as follows: 
\begin{equation}
\begin{cases}
\boldsymbol{\mu}_{k} =\boldsymbol{l}_k,\ \
\boldsymbol{\Sigma}_{k} = \operatorname{diag}\left(\frac{\boldsymbol{b}_k\circ \boldsymbol{b}_k}{4}\right) \\
\pi_k = \displaystyle{\frac{\log{s_k}}{\log{\prod_{k=1}^K s_k}}} \\
\end{cases}, \ 
\text{for}\ k=1, ..., K,
\end{equation}
where $\circ$ denotes the Hardmard product operation, and $\pi_k$ is normalized in logarithmic space to prevent the sampling probability of smaller instances from approaching zero, especially when the semantic areas vary significantly among instances.

Additionally, to address the redundant sampling issue, we incorporate constraints between the sampling PDF and existing samples, and formulate the final conditioned sampling PDF as follows:
\begin{equation}
	P(\boldsymbol{x} \mid \mathcal{X}) = 
  \begin{cases}\begin{array}{lr}
		0, & \text{if } \exists\ \boldsymbol{x}_i\in\mathcal{X}, ||\boldsymbol{x}-\boldsymbol{x}_i||_2^2<2l^2\\
		p(\boldsymbol{x}), & \text{otherwise}
	\end{array}\end{cases}
\end{equation}
where $l$ is the patch size, and $\mathcal{X}$ is an anchor set storing existing samples. With the final PDF, we randomly sample a collection of well-distributed and non-overlapping patches for image rendering and depth reconstruction. More details on the parameter initialization and the mathematical derivations of the PDF are given in our supplement.

\subsection{Loss Formulation}
A reference image can be warped to the target view using camera intrinsic parameters and differentiable grid sampling when its per-pixel depth is known. The original target image and the warped reference image should exhibit temporal photometric consistency. Furthermore, when performing volume rendering on a given frame, the rendered RGB and depth images should be respectively consistent with the original RGB image and the predicted metric depth map, thereby satisfying spatial reconstruction consistency. Therefore, we formulate a novel loss function as follows:
\begin{equation}
	\mathcal{L} = \lambda_1 \mathcal{L}_{ta} + \lambda_2 (\mathcal{L}_{rc}^{d} + \mathcal{L}_{rc}^{rgb}),
\end{equation}
where $\mathcal{L}_{ta}$ denotes the temporal alignment loss, $\mathcal{L}_{rc}^{d}$ and $\mathcal{L}_{rc}^{rgb}$ represent the reconstruction consistency losses for depth and RGB image rendering, respectively, and $\lambda_1$ and $\lambda_2$ are the weighting parameters used to balance these two types of losses.

\subsubsection{Temporal Alignment Loss}
The homogeneous coordinates $\boldsymbol{\tilde{x}}^{t}$ and $\boldsymbol{\tilde{x}}^{t+1}$ in adjacent principal frames $\boldsymbol{I}^t_0$ and $\boldsymbol{I}^{t+1}_0$ are related as follows:
\begin{equation}
	\boldsymbol{\tilde{x}}^{t+1} = \hat{\boldsymbol{D}}^t(\boldsymbol{x})\boldsymbol{K}\boldsymbol{T}\boldsymbol{K}^{-1}\boldsymbol{\tilde{x}}^t,
\end{equation}
where $\boldsymbol{T}$ denotes the relative camera pose. Therefore, we can warp $\boldsymbol{I}^{t+1}_0$ into the pixel grid of $\boldsymbol{I}^t_0$ using differentiable grid sampling, producing a synthesized image $\hat{\boldsymbol{I}}^t_0$. The temporal alignment loss, expressed as follows:
\begin{equation}
	\mathcal{L}_{ta} = \frac{1}{N}\sum_{\boldsymbol{x}}
	\boldsymbol{M}(\boldsymbol{x})\left|\boldsymbol{I}^t_0(\boldsymbol{x}) - \hat{\boldsymbol{I}}^t_0(\boldsymbol{x})\right|,
\end{equation}
can be computed to enforce photometric similarity across adjacent frames, where $\boldsymbol{M}$ represents the weight mask detailed in \cite{bian2019unsupervised} and $N$ denotes the number of valid pixels for loss computation. 

\subsubsection{Reconstruction Consistency Loss}
It is common preliminaries that $\boldsymbol{D}({\boldsymbol{x}})\boldsymbol{\tilde{x}} = \boldsymbol{K}\boldsymbol{p}$ and $||\boldsymbol{p}||_2=\hat{\boldsymbol{D}}_r(\boldsymbol{x})$. We can therefore use these relations to establish criteria for depth reconstruction consistency as follows:
\begin{equation}
	\mathcal{L}_{rc}^{d} = \frac{1}{M} \sum_{\boldsymbol{x}}
	\left|\frac{\hat{\boldsymbol{D}}_r(\boldsymbol{x})}{||\boldsymbol{K}^{-1}\boldsymbol{\tilde{x}}||_2} - \hat{\boldsymbol{D}}(\boldsymbol{x})\right|, 
\end{equation}
where $M$ denotes the number of valid pixels for loss computation. In addition, to enforce the consistency between the original and rendered RGB patches, we adopt the same rendering loss as detailed in \cite{behind2023wimbauer}:
\begin{equation}
	\mathcal{L}_{rc}^{rgb} = 
	\beta_1\operatorname{SSIM}\left(\boldsymbol{I}, \hat{\boldsymbol{I}}_{r}\right) + 
	\beta_2\left|\left|\boldsymbol{I} - \hat{\boldsymbol{I}}_{r}\right|\right|_1,
\end{equation}
where $\beta_1=0.85$ and $\beta_2=0.15$ are the empirical parameters used in \cite{behind2023wimbauer}.

\section{Experiments}
\label{sec.exp}
\subsection{Datasets, Metrics, and Implementation Details}
The 3D reconstruction performance of our proposed method is evaluated on the KITTI-360 dataset \cite{Kitti2022liao} and the KITTI Raw dataset \cite{vision2013geiger}, both providing time-stamped stereo images along with ground-truth camera poses for the evaluation of scene perception algorithms. All images are resized to the resolution of 192$\times$640 pixels, and the depth range is capped at 80m in both datasets. Following \cite{behind2023wimbauer}, we split the KITTI-360 dataset into a training set of 98,008 images, a validation set of 11,451 images, and a test set of 446 images for the 3D occupancy prediction task. We adopt the Eigen split \cite{godard2019digging} for depth estimation on the KITTI Raw dataset. Moreover,
we use the DDAD dataset \cite{guizilini20203d} to evaluate our model's zero-shot generalizability using the weights obtained on the KITTI-360 dataset. The input images, with the original resolution of 1,216$\times$1,936 pixels, are center-cropped and resized to 192$\times$640 pixels for fair comparison.

Following the experimental protocols established in previous works \cite{behind2023wimbauer, know2024li}, we quantify the 3D occupancy prediction performance of the model using the following metrics: 
scene occupancy accuracy $\text{O}_{acc}^s$,  
invisible scene accuracy $\text{IE}_{acc}^s$,
invisible scene recall $\text{IE}_{rec}^s$,
object occupancy accuracy $\text{O}_{acc}^o$,
invisible object accuracy $\text{IE}_{acc}^o$, and
invisible object recall $\text{IE}_{rec}^o$. Furthermore, we use the mean absolute relative error (Abs Rel), mean squared relative error (Sq Rel), root mean squared error (RMSE), root mean squared log error (RMSE log), and accuracy under certain thresholds ($\delta_i < 1.25^i, i = 1, 2, 3$) to quantify the model's monocular depth estimation performance.

\begin{table}
	\centering
	\settablefont
	
	\begin{tabular}{l|ccc}
		\toprule
		Method 
		&$\text{O}_{acc}^s\uparrow$ 
		&$\text{IE}_{acc}^s\uparrow$
		&$\text{IE}_{rec}^s\uparrow$\\ \hline
		Monodepth2 \cite{godard2019digging}     & 0.90 & N/A  & N/A   \\
		Monodepth2 + 4m							& 0.90 & 0.59 & 0.66  \\
		PixelNeRF \cite{yu2021pixelnerf}        & 0.89 & 0.62 & 0.60  \\
		BTS \cite{behind2023wimbauer}           & 0.92 & 0.69 & 0.64  \\
		KYN \cite{know2024li}                   & 0.92 & 0.70 & 0.66  \\ \hline
		\textbf{ViPOcc (Ours)}					& \textbf{0.93} & \textbf{0.71} & \textbf{0.69}\\
		\bottomrule
	\end{tabular}
	\caption{Comparison of scene reconstruction performance on the KITTI-360 dataset.} 
	\label{tb.scene_recon}
\end{table}

\begin{table}
	\centering
	\settablefont
	
	\begin{tabular}{l|ccc}
		\toprule
		Method 
		&$\text{O}_{acc}^o\uparrow$ 
		&$\text{IE}_{acc}^o\uparrow$
		&$\text{IE}_{rec}^o\uparrow$\\ \hline
		Monodepth2 \cite{godard2019digging}              	& 0.69 & N/A  & N/A  \\
		Monodepth2 + 4m							            & 0.70 & 0.53 & 0.52 \\
		PixelNeRF \cite{yu2021pixelnerf}            		& 0.67 & 0.53 & 0.49 \\
		BTS \cite{behind2023wimbauer}           			& 0.79 & 0.69 & 0.60 \\
		KYN \cite{know2024li}                   			& 0.79 & 0.69 & 0.61 \\ \hline
		\textbf{ViPOcc (Ours)}							 	& \textbf{0.79} & \textbf{0.69} & \textbf{0.64} \\
		\bottomrule
		
	\end{tabular}
	\caption{Comparison of object reconstruction performance on the KITTI-360 dataset.}
	\label{tb.obj_recon}
\end{table}

The proposed method is trained on an NVIDIA RTX 4090 GPU using the Adam optimizer for 25 epochs, with an initial learning rate of 1e-4, which is reduced by a factor of 10 during the final 10 epochs. We use BTS \cite{behind2023wimbauer} as our baseline network and adopt the metric depth predictions from Depth Anything V2 \cite{yang2024depth} as pseudo depth. We use Grounded-SAM \cite{ren2024grounded} to generate instance-level semantic masks and bounding boxes.

\subsection{Comparisons with State-of-The-Art Methods}
\subsubsection{3D Occupancy Prediction}

Following the experimental protocols detailed in the study \cite{behind2023wimbauer}, we compare ViPOcc with a representative self-supervised monocular depth estimation network Monodepth2 \cite{godard2019digging} and other NeRF-based SoTA methods in terms of 3D occupancy prediction performance, as presented in Tables \ref{tb.scene_recon} and \ref{tb.obj_recon}. Specifically, when evaluating Monodepth2's 3D occupancy prediction performance, all points behind visible pixels in the image are considered occupied. This is primarily due to the infeasibility of inferring the true 3D geometry of points that are invisible in the image. Furthermore, we also follow prior studies \cite{know2024li, behind2023wimbauer} to quantify the model's performance by considering points within a distance of up to 4m from visible points as occupied. 

It can be observed that ViPOcc achieves SoTA performance across all metrics in 3D occupancy prediction for both scene and object reconstruction. Notably, $\text{O}_{acc}^s$, $\text{O}_{rec}^s$, and $\text{O}_{rec}^o$ increase by 1.1-3.4\%, 4.6-15.0\%, and 4.9-30.6\%, respectively. It is also worth noting that Monodepth2 + 4m can deliver competitive performance in $\text{O}_{acc}^s$ and $\text{O}_{acc}^o$. However, it relies on hand-crafted criteria rather than directly learning the 3D structure from a single view \cite{know2024li}.

\begin{figure}[!t]
	\centering
	\includegraphics[width=0.99\linewidth]{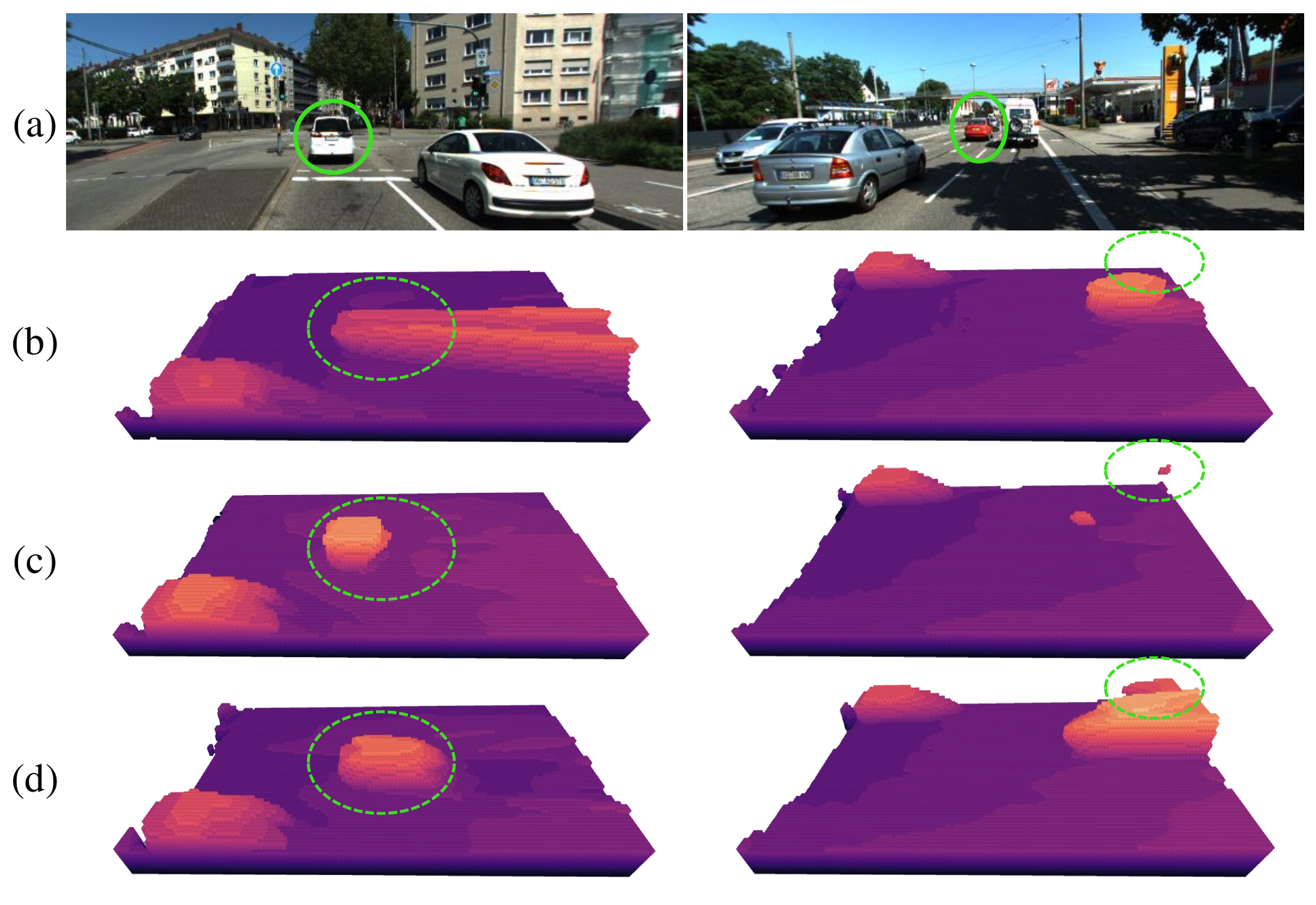}
	\caption{Qualitative comparison of 3D occupancy prediction on the KITTI-360 dataset: (a) input RGB images; (b) BTS results; (c) KYN results; (d) our results.  A darker voxel color indicates a lower altitude.}
	\label{fig.results_occ}
\end{figure}

Qualitative comparisons are presented in Fig. \ref{fig.results_occ}, where the predicted occupancy grids are viewed from the right side of the scene. It is evident that our method significantly outperforms both BTS and KYN in 3D geometry reconstruction, particularly for crucial instances, and effectively reduces trailing effects. These results demonstrate the efficacy of ViPOcc in reasoning about occluded regions against inherent ambiguities.

\begin{table*}[!t]
	\centering
	\settablefont
	
	\begin{tabular}{l|cccc|ccc}
		\toprule
		Method & Abs Rel $\downarrow$ & Sq Rel $\downarrow$ & RMSE $\downarrow$ & RMSE log $\downarrow$ & $\delta <1.25$ $\uparrow$&  $\delta <1.25^2$ $\uparrow$ & $\delta <1.25^3$ $\uparrow$ \\
		\hline
		Monodepth2 \cite{godard2019digging}		& 0.106 & 0.818 & 4.750 & 0.196 & 0.874 & 0.957 & 0.979 \\
		SwinDepth \cite{shim2023swindepth}		& 0.106 & 0.739 & 4.510 & 0.182 & 0.890 & 0.964 & \textbf{0.984} \\
		Lite-Mono \cite{zhang2023lite}			& 0.107 & 0.765 & 4.561 & 0.183 & 0.886 & 0.963 & 0.983 \\	
		Dynamo-Depth \cite{sun2024dynamo}		& 0.112 & 0.758 & 4.505 & 0.183 & 0.873 & 0.959 & 0.984 \\
		BTS \cite{behind2023wimbauer}			& 0.102	& 0.755	&\textbf{4.409}	& 0.188	&0.882	&0.961	& 0.982 \\
		MVBTS \cite{han2024boosting}			&0.105	&0.757	&4.501	&0.193	&0.873	&0.957	&0.981  \\
		KDBTS \cite{han2024boosting}			&0.105	&0.761	&4.498	&0.193	&0.873	&0.957	&0.981  \\
		\hline
		\textbf{ViPOcc (Ours)}                  &\textbf{0.096}  &\textbf{0.652}  &4.507  &\textbf{0.179}  &\textbf{0.890}   &\textbf{0.964}  &0.983\\  
		\bottomrule
		
	\end{tabular}
	\caption{Comparison of depth estimation performance on the KITTI Raw dataset using the Eigen split.}
	\label{tb.depth_kitti_raw}
\end{table*}

\begin{figure}[!t]
	\centering
	\includegraphics[width=0.99\linewidth]{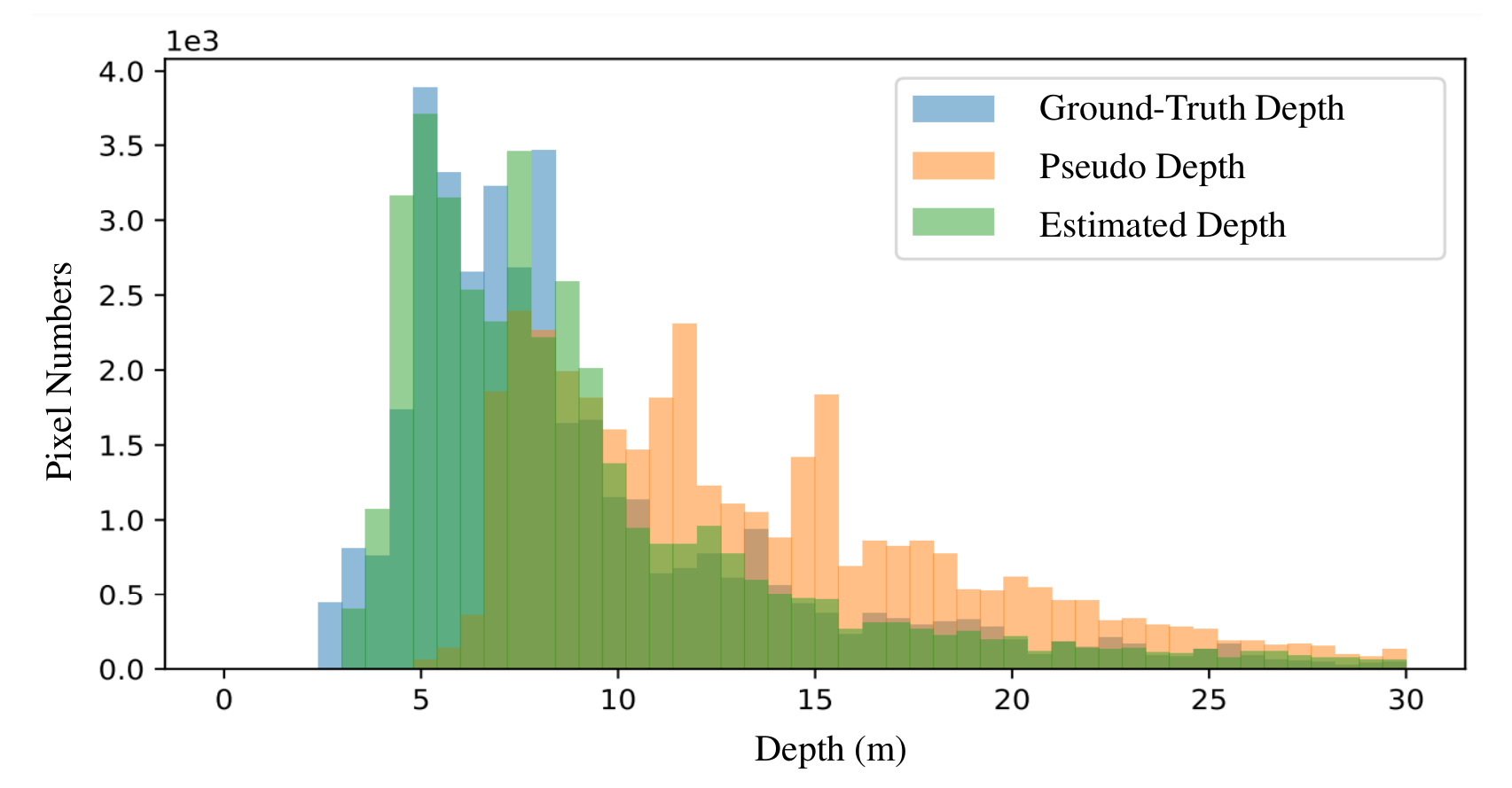}
	\caption{Depth distribution comparison.}
	\label{fig.depth_distri}
\end{figure}

\begin{figure}[!t]
	\centering
	\includegraphics[width=0.99\linewidth]{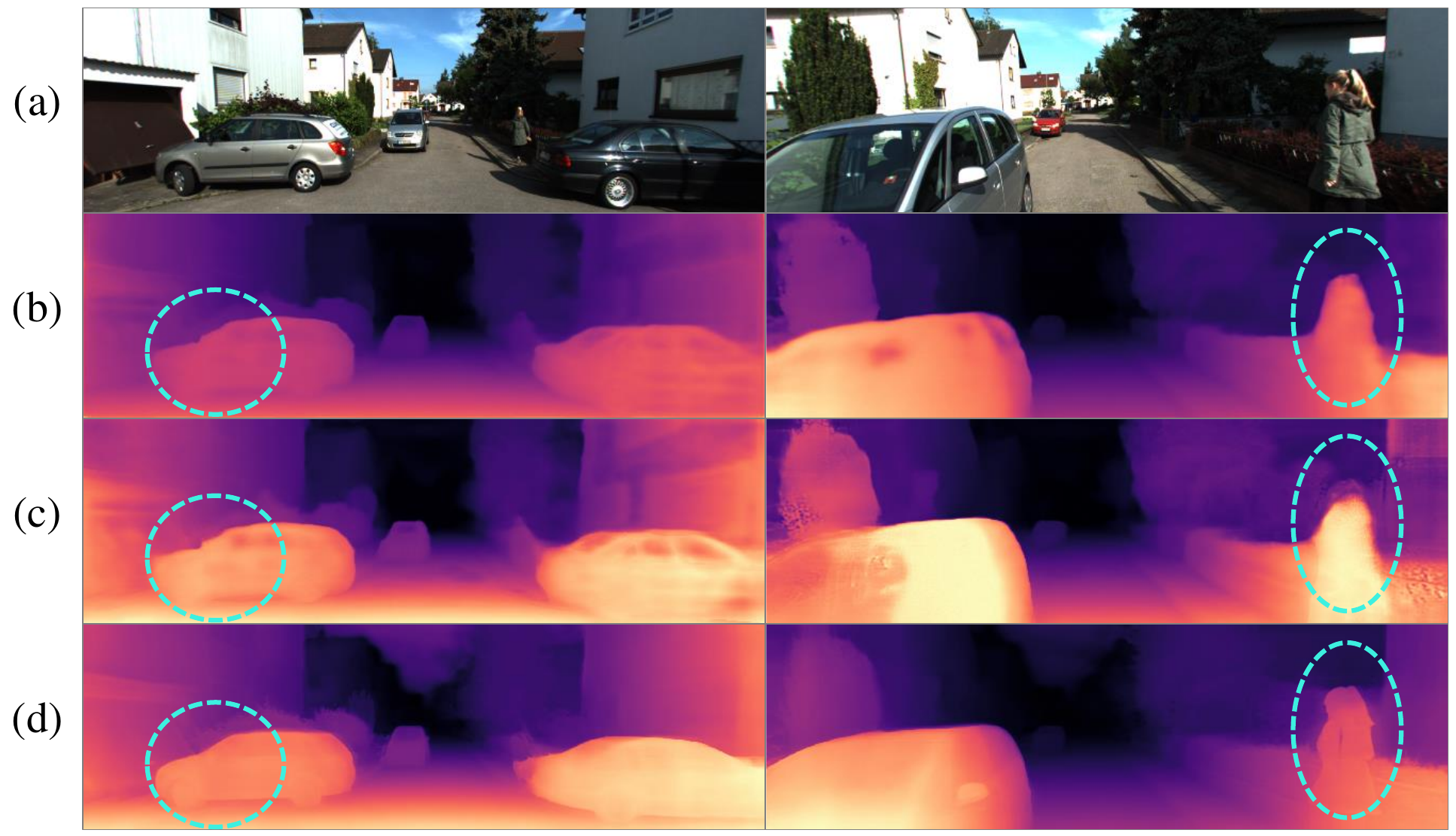}
	\caption{Comparison of metric depth estimation on the KITTI-360 dataset: (a) input RGB images; (b) BTS results; (c) KYN results; (d) our results. }
	\label{fig.results_depth_kitti360}
\end{figure}

\begin{table}
	\centering
	\settablefont

	\begin{tabular}{lccc}
		\toprule
		Method   &Abs Rel &RMSE log & $\alpha<1.25$\\
		\hline
		Pseudo depth (no scaling)		&0.586	&0.477	&0.071\\
		Pseudo depth (median scaling)	&0.142	&0.209	&0.832\\
		BTS \cite{behind2023wimbauer}    	&0.103	&0.194	&\textbf{0.891}\\
		KYN	\cite{know2024li} 				&0.107	&0.197	&0.880\\
		\hline
		\textbf{ViPOcc (Ours)} 				&\textbf{0.097}	&\textbf{0.188}	&0.886\\
		\bottomrule
	\end{tabular}
	\caption{Comparison of metric depth estimation performance on the KITTI-360 dataset.}
	\label{tb.depth_kitti360}
\end{table}

\subsubsection{Metric Depth Estimation}

Table \ref{tb.depth_kitti360} shows the comparison of metric depth estimation performance among VFM, previous SoTA methods, and our proposed ViPOcc on the KITTI-360 dataset. It is evident that the depth predictions from Depth Anything V2 \cite{yang2024depthv2} are unsatisfactory, regardless of whether median scaling is used to align the depth distribution. Notably, ViPOcc demonstrates superior performance compared to existing NeRF-based unsupervised methods, achieving a decrease of 5.8\% in Abs Rel. Moreover, we compare depth distributions among ground-truth depth, pseudo depth, and our predictions. As illustrated in Fig. \ref{fig.depth_distri}, significant discrepancies among these distributions not only underscore the infeasibility of directly using pseudo depth to generate supervisory signals, but also validate the effectiveness of our proposed inverse depth alignment module in refining depth.

As presented in Fig. \ref{fig.results_depth_kitti360}, the qualitative comparison with prior SoTA methods on the KITTI-360 dataset demonstrates ViPOcc's exceptional metric depth estimation performance. Our method exhibits superior depth consistency in continuous regions, as shown on the vehicle’s glass, and preserves clear object boundaries, as shown on the pedestrian. These improvements can be attributed to the spatial reconstruction consistency constraint we incorporated between rendered and predicted depth maps, which also preserves the local differential properties of VFM predictions to enable fine-grained depth estimation.

Moreover, as presented in Table \ref{tb.depth_kitti_raw}, ViPOcc also demonstrates superior depth estimation performance compared to all existing self-supervised methods on the KITTI Raw dataset. Specifically, it achieves a decrease of 5.9\% in Abs Rel and  11.8\% in Sq Rel compared to previous SoTA approaches. Surprisingly, ViPOcc significantly outperforms its counterparts trained with the same NeRF-based architectures, such as BTS \cite{behind2023wimbauer}, MVBTS \cite{han2024boosting}, and KDBTS \cite{han2024boosting}. It achieves an average error reduction of 9.5\% in Abs Rel and an average performance gain of 1.3\% in $\delta_1$. These experimental results underscore the effectiveness of our proposed ViPOcc framework for monocular depth estimation across different scenarios with distinct experimental setups.

\begin{table}[!t]
	\centering
	\settablefont

	\begin{tabular}{cl|ccc}
		\toprule
		\multicolumn{2}{c|}{Configuration}   & $\text{O}_{acc}^s\uparrow$ & $\text{IE}_{acc}^s\uparrow$ & $\text{IE}_{rec}^s\uparrow$ \\  
		\hline 
		\multicolumn{2}{c|}{Baseline \cite{behind2023wimbauer}} & 0.91 & 0.65 & 0.64 \\ 
		\hline
		\multicolumn{1}{c}{\multirow{3}{*}{\makecell{Depth\\estimation}}} 
		&+ Depth 		&0.91 &0.64 &0.64	 \\
		&+ Pseudo depth		&0.86 &0.60 &0.61	 \\
		&+ Inverse pseudo depth		&0.92 &0.65 &0.66	 \\
		\hline
		\multicolumn{1}{c}{\multirow{2}{*}{\makecell{Ray\\sampling}}} 
		&+ Random sampler	& 0.91 & 0.65 & 0.64\\
		&+ SNOG sampler		& 0.92 & 0.67 & 0.67\\
		\hline
		\multicolumn{1}{c}{\multirow{3}{*}{Loss}} 
		&+ $\mathcal{L}_{ta}$	&0.91 &0.66 &0.64 \\
		&+ $\mathcal{L}_{rc}$	&0.89 &0.64 &0.60 \\
		&+ $\mathcal{L}_{ta}$, $\mathcal{L}_{rc}$ 	&0.92 &0.69 &0.67\\
		\hline
		\multicolumn{2}{c|}{Full implementation} &\textbf{0.93} &\textbf{0.71} &\textbf{0.69}	\\ 
		\bottomrule
	\end{tabular}
	\caption{Ablation studies of ViPOcc inner designs on the KITTI-360 dataset.}
	\label{tb.ablation}
\end{table}

\subsection{Ablation Studies}

We validate the rationality and efficacy of ViPOcc through extensive ablation studies, specifically focusing on depth estimation methods, ray sampling strategies, and loss function designs, as presented in Table \ref{tb.ablation}. 

We first adopt an individual depth prediction branch without VFM's visual priors incorporated for depth estimation, resulting in performance similar to that of the baseline. We attribute this phenomenon to a performance bottleneck within the depth prediction network, due to its estimations not being sufficiently accurate, which in turn limits improvements in 3D occupancy prediction. We then investigate the effectiveness of aligning VFM's depth priors based on depth residuals. As discussed earlier, neural networks struggle to converge or maintain accuracy when fitting depth residuals, which typically exhibit a large range of numerical variations. Consequently, a drastic performance drop occurs, falling within our expectations. When employing our proposed inverse depth alignment module, a notable performance improvement in 3D occupancy prediction is achieved, demonstrating its effectiveness.

Moreover, as observed, the SNOG sampler leads to improved performance, particularly in invisible scene accuracy and recall, which increase by 3.1-4.7\%. This validates the effectiveness of our proposed ray sampling strategy. Additional comparisons between random and SNOG samplers regarding their efficiency are provided in our supplement.

In addition, it is evident that relying solely on temporal alignment loss yields limited performance improvements, whereas using only the reconstruction consistency loss actually degrades the framework's performance. However, combining both losses significantly enhances 3D occupancy prediction performance, leading to an increase of approximately 4.7-6.2\% in invisible scene accuracy and recall.

\subsection{Zero-Shot Depth Estimation}
To further evaluate the generalizability of ViPOcc, we conduct a zero-shot test on the DDAD dataset \cite{guizilini20203d} using the pre-trained weights obtained from the KITTI-360 dataset. As presented in Table \ref{tb.depth_ddad}, ViPOcc outperforms other SoTA methods in zero-shot depth estimation, demonstrating its exceptional generalizability.

\begin{table}
	\centering
	\settablefont

	\begin{tabular}{lccc}
		\toprule
		Method   &Abs Rel &RMSE log & $\alpha<1.25$\\
		\hline
		BTS \cite{behind2023wimbauer}    	&0.182 &0.290 &0.746\\ 
		KYN	\cite{know2024li} 				&0.190 &0.286 &0.749\\
		\hline
		\textbf{ViPOcc (Ours)} 				&\textbf{0.175}	&\textbf{0.282}	&\textbf{0.749}\\
		\bottomrule
	\end{tabular}
	\caption{Zero-shot depth estimation performance comparison on the DDAD dataset.}
	\label{tb.depth_ddad}
    \vspace{-1em}
\end{table}

\section{Conclusion}
This paper introduced ViPOcc, a novel framework that effectively leverages VFM's visual priors for single-view 3D occupancy prediction.
ViPOcc consists of two coupled branches: one estimates highly accurate metric depth by aligning the inverse depth output from Depth Anything V2, while the other one predicts 3D occupancy with a Grounded-SAM-guided Gaussian mixture sampler incorporated to achieve efficient and instance-aware ray sampling. These two branches are effectively coupled through a temporal photometric alignment loss and a spatial geometric consistency loss. Extensive experiments and comprehensive analyses validate the effectiveness of our novel contributions and the superior performance of ViPOcc compared to previous SoTA methods. In the future, we aim to achieve a tighter coupling of these two branches and develop a more lightweight 3D occupancy prediction framework.


{
\small
\bibliographystyle{ieeenat_fullname}
\bibliography{ref}
}

\clearpage

\appendix
\setcounter{equation}{0}
\renewcommand\theequation{A.\arabic{equation}}

\section{Additional Methodological Details}
\subsection{Framework Architecture}

For the parallel encoders, we utilize two ResNet-50 \cite{he2016deep} backbones, each pre-trained on the ImageNet dataset \cite{deng2009imagenet}. These backbones share the same architecture and initialization process but differ in their self-supervisory signals. For the task-specific decoders, we adopt the architecture of Monodepth2 \cite{godard2019digging} with modifications to the final layers. As detailed in Table \ref{tb.arch_decoder}, the output features from both decoders, $\boldsymbol{F}_r$ and $\boldsymbol{F}_s$, are upscaled to a resolution of $192\times640$ pixels. These features are then utilized for residual inverse depth estimation and dense 3D scene reconstruction.

\subsection{SNOG Sampler}
This subsection presents preliminaries on the Gaussian distribution and implementation details of the proposed SNOG sampler.

\subsubsection{Gaussian Distribution}
In probability theory, the normal or Gaussian distribution is a type of continuous probability distribution characterized by the following probability density function (PDF):
\begin{equation}
	f(x)=\frac{1}{\sqrt{2\pi\sigma^2}}e^{-\frac{(x-\mu)^2}{2\sigma^2}},
\end{equation}
where $\mu$ denotes the mean or expectation of the distribution, and $\sigma$ represents the variance. The associated error function $\operatorname{erf}(x)$ calculates the probability that a random variable with a normal distribution, having a mean of 0 and a variance of $\frac{1}{2}$, falls within the range $[-x,x]$, expressed as follows:
\begin{equation}
	\operatorname{erf}(x)={\frac{1}{\sqrt{\pi}}}
	\int_{-x}^{x}e^{-t^{2}}dt={\frac{2}{\sqrt{\pi}}}
	\int_{0}^{x}e^{-t^{2}}dt.
\end{equation}
Moreover, the probability that a random variable with a Gaussian distribution lies between $\mu-n\sigma$ and $\mu+n\sigma$ is given by $\operatorname{erf}(\frac{n}{\sqrt{2}})$. Therefore, for an arbitrary normal distribution, values within two standard deviations from the mean approximately account for 95.45\% of the distribution.

\begin{table}[t!]
	\centering
	\settablefont

	\begin{tabular}{lcccc}
		\toprule
		\multicolumn{5}{c}{\multirow{1}*{Task-Specific Decoders}}           \\ \hline
		Layer            & KS & Channels                & Resolution        & Activation \\ \hline
		Conv2d           & 3  & 512$\rightarrow$256 & $(6\times20) $    & ELU        \\
		UpFusion         & 3  & 256                 & $(12\times40)$    & ELU        \\ \hline
		Conv2d           & 3  & 128                 & $(12\times40)$    & ELU        \\
		UpFusion         & 3  & 128                 & $(24\times80)$    & ELU        \\ \hline
		Conv2d           & 3  & 64                  & $(24\times80)$    & ELU        \\
		UpFusion         & 3  & 64                  & $(48\times160)$   & ELU        \\ \hline
		Conv2d           & 3  & 32                  & $(48\times160)$   & ELU        \\
		UpFusion         & 3  & 32                  & $(96\times320)$   & ELU        \\ \hline
		Conv2d           & 3  & 16                  & $(96\times320)$   & ELU        \\
		UpFusion         & 3  & 16                  & $(192\times640)$  & ELU        \\ \hline
		Conv2d $\rightarrow$ $\boldsymbol{F}_r$     & 3  & 64   & $(192\times640)$  & -          \\
		Conv2d $\rightarrow$ $\boldsymbol{F}_s$     & 1  & 64   & $(192\times640)$  & -          \\         
		\bottomrule
	\end{tabular}
	\caption{Network architecture of task-specific decoders. ``KS'' represents the kernel size, and the ``UpFusion'' layer represents a combination of an upsampling layer, a skip connection layer, and a convolutional layer.}
	\label{tb.arch_decoder}
\end{table}
\subsubsection{Derivation of the Sampling PDF}
Specifically, we utilize processed ground-truth labels from the Cityscapes semantic segmentation task \cite{cordts2016cityscapes} as prompts for Grounding DINO \cite{liu2023grounding}. As shown in Table \ref{tb.prompt}, to avoid excessively detected bounding boxes and fragmented segmentation results, we exclude uncommon phrases, such as ``bridge'' and ``tunnel'', and merge semantically similar phrases, such as ``truck'', ``bus'', and ``caravan'' provided in the original Cityscapes dataset. Furthermore, these phrases are categorized for different sampling strategies based on their physical properties. For phrases typically associated with large areas, such as ``sky'', ``road'', ``vegetation'', and ``building'', we employ a 2D uniform sampling strategy to ensure evenly allocated attention within these areas. For critical regions, such as ``car'' and ``pedestrian'', we adopt a Gaussian mixture sampling strategy to allow for a higher density of sampling points, thereby enhancing fine-grained 3D reconstruction performance. Moreover, we employ a uniform sampling strategy for background regions, namely ``unlabeled''. Therefore, if Grounding DINO detects no objects in an input image, our SNOG sampler automatically defaults to the random uniform sampler used in previous state-of-the-art methods \cite{behind2023wimbauer, know2024li}.

As described in the main paper, we utilize SAM \cite{kirillov2023segment} to obtain the metadata $\mathcal{M}_k=\{\boldsymbol{l}_k, \boldsymbol{b}_k, s_k\}$ for the $k$-th instance.
Therefore, the instance-level PDF $p_k(\boldsymbol{x}): \mathbb{R}^{H\times W} \rightarrow [0,1)$ can be formulated as follows:
\begin{equation}
	p_k(\boldsymbol{x}) = 
	\begin{cases}\begin{array}{lr}
			\mathcal{N}\left(\boldsymbol{x} \mid \boldsymbol{\mu}_{k},\boldsymbol{\Sigma}_{k}\right), & \text{if } \mathcal{M}_k \in \mathcal{G}\\
			\mathcal{U}\left(\boldsymbol{x} \mid s_k\right), & \text{otherwise}
	\end{array}\end{cases}
\end{equation}
where $\mathcal{G}$ denotes the Gaussian sampling metadata set.
Furthermore, to derive a plausible sampling PDF for the entire scene, we employ a mixture sampling model to combine all instance-level PDFs, yielding the scene-level PDF $p(\boldsymbol{x})$ as follows:
\begin{equation}
	\begin{aligned}
		p(\boldsymbol{x})&=\sum_{i=1}^{N}p_k(\boldsymbol{x})\\
		&=\sum_{k=1}^{K}\pi_{k}\mathcal{N}\left(\boldsymbol{x} \mid \boldsymbol{\mu}_{k}, \boldsymbol{\Sigma}_{k}\right)
		+\sum_{j=K+1}^{N}{\pi_{j}\mathcal{U}\left(\boldsymbol{x} \mid s_k\right)} \\
		&=(1-\gamma)\sum_{k=1}^{K} \pi_{k} \mathcal{N}\left(\boldsymbol{x} \mid \boldsymbol{\mu}_{k}, \boldsymbol{\Sigma}_{k}\right) + 
		\gamma\mathcal{U}\left(\boldsymbol{x} \mid s\right), 
	\end{aligned}
	\label{eq.p(x)}
\end{equation}
where $\gamma$ denotes the background sampling ratio, $N$ and $K$ denote the total number of instances and the number of Gaussian sampling instances, respectively, and $s=\sum{s_k}$ represents the total area of uniformly sampled regions. Based on (\ref{eq.p(x)}) in the supplement, we further formulate the conditioned sampling PDF $P(\boldsymbol{x} | \mathcal{X})$ (discussed in the main paper) for efficient and instance-aware ray sampling. Qualitative visualizations of our SNOG sampler are presented in Fig. \ref{fig.results_sampler} in the supplement.

\begin{table}[t!]
	\centering
	\settablefont

	\begin{tabular}{c|c}
		\toprule
		Prompt		&Sampling strategy \\ \hline
		Road		&$\mathcal{U}$\\
		Building	&$\mathcal{U}$\\
		Vegetation	&$\mathcal{U}$\\
		Sky			&$\mathcal{U}$\\
		Car			&$\mathcal{N}$\\
		Pedestrian	&$\mathcal{N}$\\
		Unlabeled	&$\mathcal{U}$\\
		\bottomrule
	\end{tabular}
	\caption{Grounding DINO prompts and their corresponding sampling strategies. $\mathcal{N}$ denotes the Gaussian sampling strategy, and $\mathcal{U}$ denotes the uniform sampling strategy.}
	\label{tb.prompt}
\end{table}

\section{Technical Details}
\subsection{Implementation Details}
Our implementation is based on the official code repository published by \cite{behind2023wimbauer}. 
Additionally, we are inspired by the repository from \cite{know2024li} regarding the evaluation of 3D occupancy prediction and visualization of density fields.

\subsubsection{Acquisition of Visual Priors}
In practice, visual priors from VFMs are pre-generated to improve training efficiency. Specifically, within the depth prediction branch, we employ a Depth Anything V2 \cite{yang2024depthv2} to produce metric pseudo depth maps, which are stored as 32-bit float ndarrays. In the SNOG sampler, we use a Grounded-SAM \cite{ren2024grounded} to pre-generate instance-level bounding boxes, semantic masks, and areas, as well as to compute the 2D coordinates of sampling anchors in advance. Both the pseudo depth maps and sampling anchors are directly loaded to accelerate the training process.

\begin{figure*}[!t]
	\centering
	\includegraphics[width=0.99\linewidth]{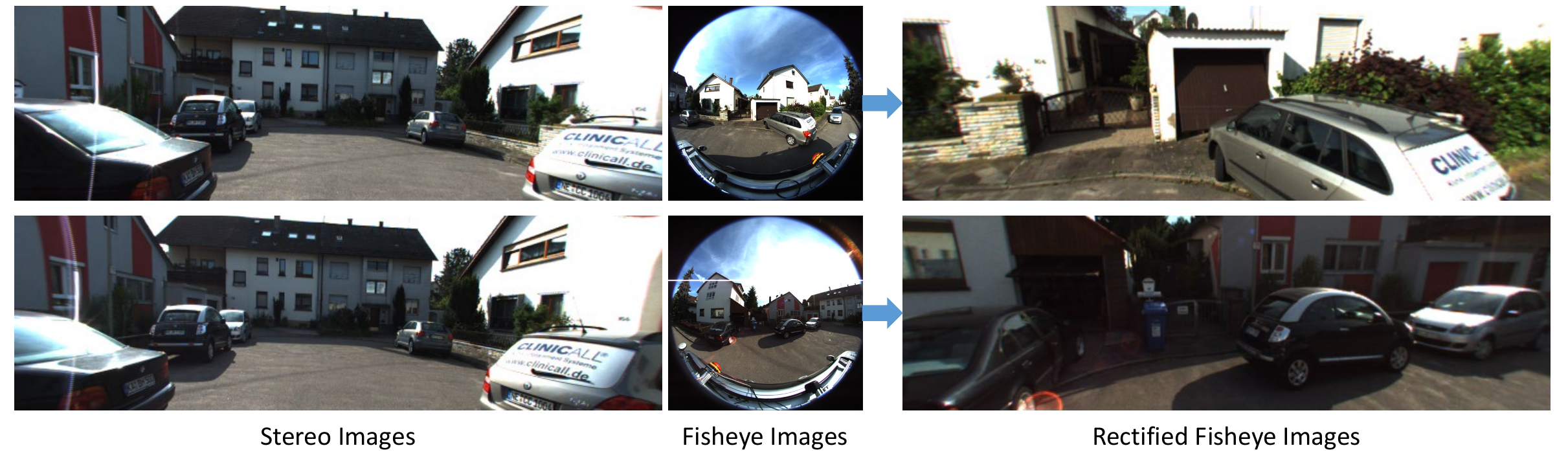}
	\caption{An illustration of stereo and fisheye images from the KITTI-360 dataset. The fisheye images are resampled using a virtual perspective camera, which shares the same intrinsic parameters as the perspective forward-facing cameras.}
	\label{fig.kitti360_input}
\end{figure*}

\subsubsection{Evaluation Protocols}
During the evaluation of 3D occupancy prediction, samples are evenly distributed within a cuboid space $\mathbb{R}^{h\times w\times d}$ relative to the camera, with dimensions $w = [-4, 4]$m, $h = [-1, 0]$m, and $d = [4, 20]$m. For each voxel $\boldsymbol{p}$, the predicted density $\sigma_{\boldsymbol{p}}$ is discretized into a binary occupancy score $o_{\boldsymbol{p}}\in\{0,1\}$ with a threshold $\tau=0.5$. 

During the evaluation of metric depth estimation, our method can produce an estimated depth map through either the depth estimation branch or the 3D occupancy prediction branch. Although their metrics are numerically close, due to the use of the reconstruction consistency loss (detailed in equation (10) in the main paper), repeated experiments have shown that the metrics of the rendered depth $\hat{\boldsymbol{D}}_r$ are often slightly superior to those of the predicted depth $\hat{\boldsymbol{D}}$. However, $\hat{\boldsymbol{D}}$ provides much better visual quality compared to $\hat{\boldsymbol{D}}_r$. This discrepancy can be attributed to the following two reasons:
\begin{itemize}
	\item The 3D occupancy prediction branch employs NeRF-based volume rendering for metric depth recovery, which provides superior global structure modeling capabilities compared to methods that directly estimate depth from 2D images.
	\item Our proposed inverse depth alignment module effectively eliminates domain discrepancies while preserving the sharpness and detail of the original depth priors from the VFM.
\end{itemize}
Therefore, we use $\hat{\boldsymbol{D}}_r$ for evaluation and $\hat{\boldsymbol{D}}$ for visualization in all experiments.

\subsection{Dataset Configuration}
\subsubsection{KITTI-360}
As illustrated in Fig. \ref{fig.kitti360_input}, the KITTI-360 dataset \cite{Kitti2022liao} provides sequential frames using two stereo cameras and two fisheye cameras. Following \cite{behind2023wimbauer}, the fisheye images are resampled using a virtual perspective camera, which shares the same intrinsic parameters as the perspective forward-facing cameras. To maximize the overlap of the viewing frustums with the forward-facing cameras, we tilt the virtual camera downward by 15$^\circ$. However, the fisheye and forward-facing cameras from the same timestamp typically do not overlap in their visible areas. Therefore, to enhance rendering quality, we offset the fisheye cameras by 10 timestamps, optimizing the overlap of their frustums.

Following \cite{know2024li}, we accumulate 3D LiDAR sweeps over a period of 300 timestamps to accurately ``carve out" the 3D scene geometry. Points that remain invisible in all LiDAR sweeps are considered occupied, while those visible in any sweep are considered as unoccupied.

\subsubsection{KITTI Raw}
We train ViPOcc for 50 epochs on the KITTI Raw dataset \cite{vision2013geiger} using the Eigen split \cite{godard2019digging}. The stereo image pairs with adjacent timestamps are fed into the network for image rendering and photometric reconstruction. Due to the absence of 3D ground-truth labels, we only report the depth estimation metrics on the KITTI Raw dataset. 

\subsubsection{DDAD}
We evaluate the generalizability of our model using the official test set of the DDAD dataset \cite{guizilini20203d}. To perform zero-shot evaluation, we directly utilize the official weights for the KITTI-360 dataset provided by BTS \cite{behind2023wimbauer} and KYN \cite{know2024li}. Due to the resolution discrepancy between the KITTI-360 dataset (192$\times$640 pixels) and the DDAD dataset (1,216$\times$1,936 pixels), we apply a vertical crop to align the fields of view. Additionally, to minimize differences in depth scale, we employ median scaling, adjusting the predicted depths using the ratio of median ground-truth depth to median predicted depth.

\begin{figure*}[!t]
	\centering
	\includegraphics[width=0.99\linewidth]{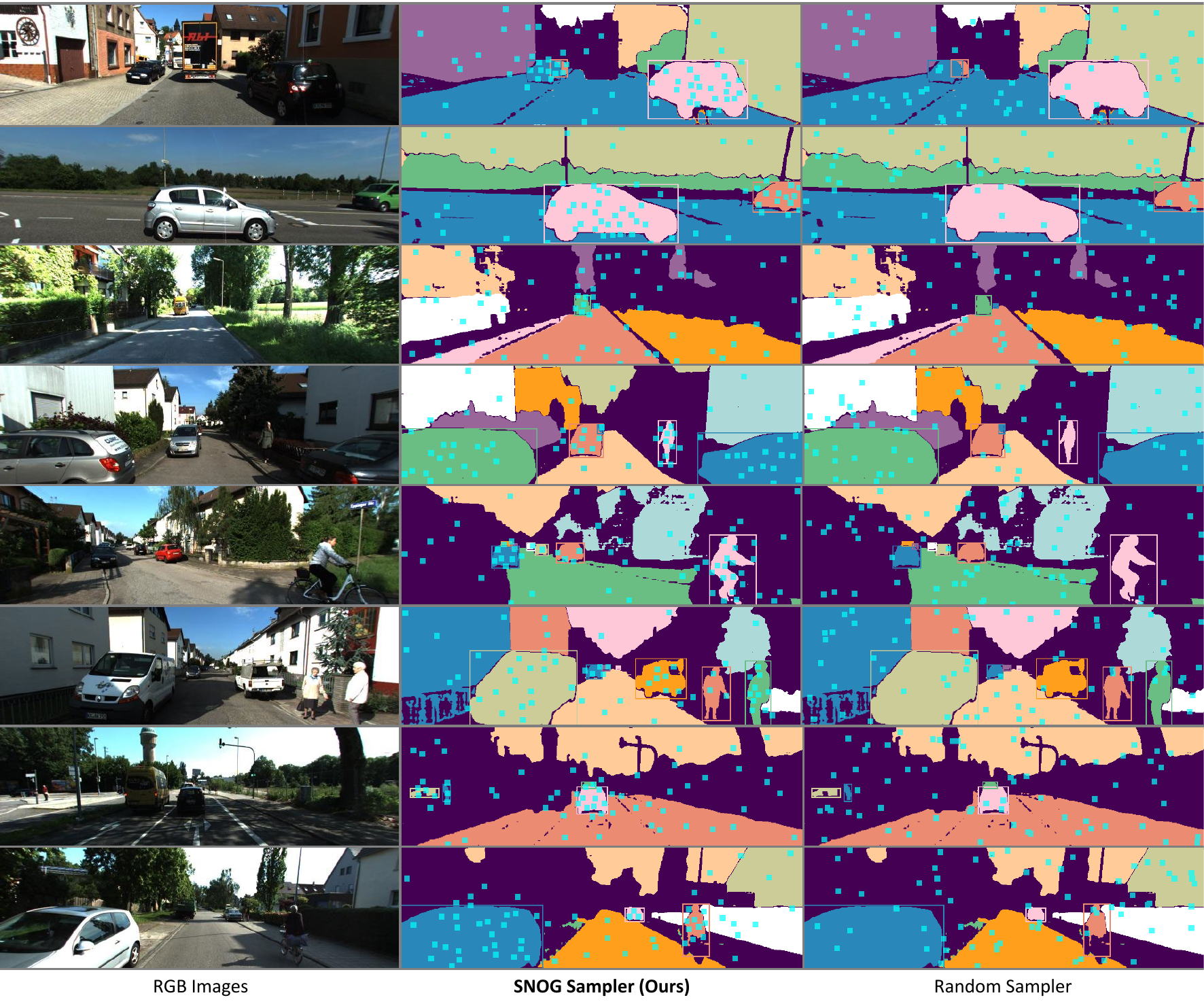}
	\caption{Comparison of sampled patches achieved using our proposed SNOG sampler and the random sampler.}
	\label{fig.results_sampler}
\end{figure*}

\begin{figure*}[!t]
	\centering
	\includegraphics[width=0.99\linewidth]{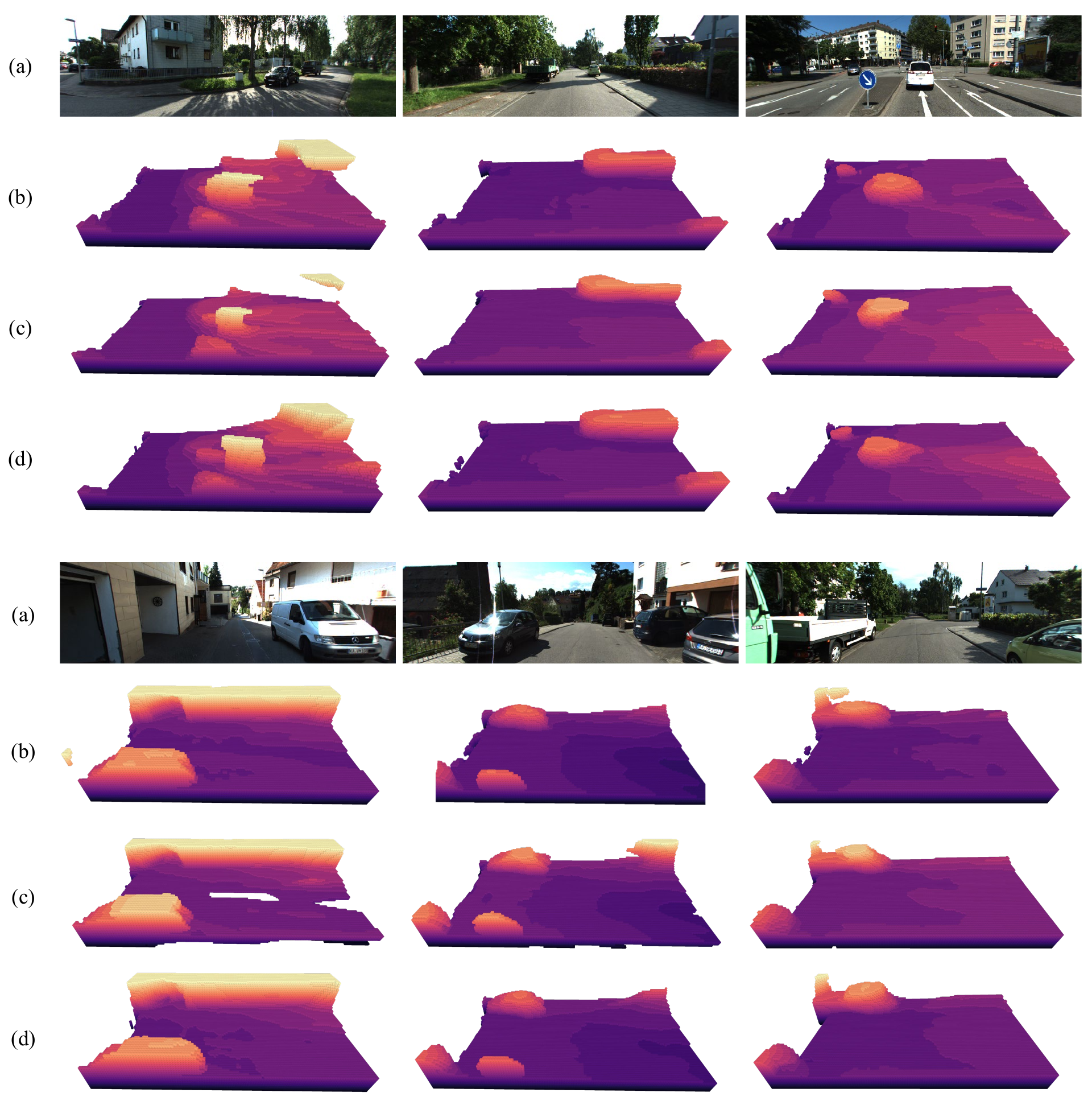}
	\caption{Comparison of 3D occupancy prediction on the KITTI-360 dataset: (a) input RGB images; (b) BTS \cite{behind2023wimbauer} results; (c) KYN \cite{know2024li} results; (d) our results.}
	\label{fig.results_occ_supp}
\end{figure*}

\begin{figure*}[!t]
	\centering
	\includegraphics[width=0.99\linewidth]{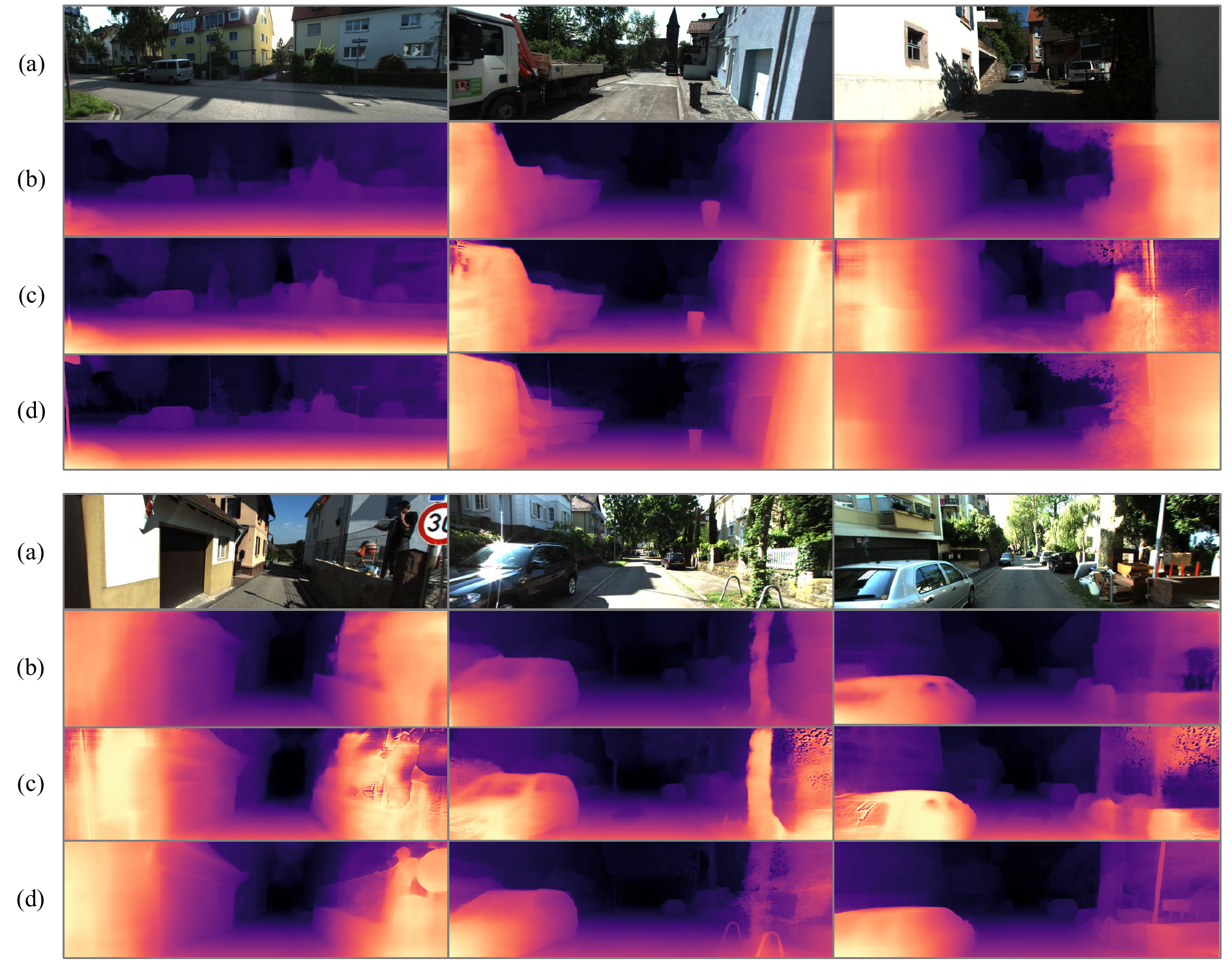}
	\caption{Comparison of metric depth estimation on the KITTI-360 dataset:  (a) input RGB images; (b) BTS \cite{behind2023wimbauer} results; (c) KYN \cite{know2024li} results; (d) our results.}
	\label{fig.results_depth}
\end{figure*}

\section{Additional Experimental Results}
We conduct additional experiments to compare network inference speed and training parameters, evaluate the computational efficiency of ray samplers, compare the performance of 3D occupancy prediction and metric depth estimation, and select hyper-parameters for loss functions.

\subsection{Network Parameters and Inference Speed}
We provide network parameters and inference speed comparisons between prior works and ViPOcc. As shown in Table \ref{tb.tp_is}, KYN \cite{know2024li} has much more training parameters due to the inclusion of a large language model, while our method maintains comparable parameters to BTS \cite{behind2023wimbauer}. The 3D occupancy and depth inference speed are tested on an NVIDIA RTX 4090D GPU using the KITTI-360 test set containing 446 images. RGB images are resized to a resolution of $192\times640$ pixels before inference. ViPOcc achieves comparable 3D occupancy inference speed to BTS, while providing significantly faster depth inference compared to the other two methods. This improvement is due to our efficient depth estimation branch, which avoids the volume rendering process during depth estimation.

\begin{table}
	\centering
	\settablefont
	\begin{tabular}{l|ccc}
		\toprule
		Method 
		&TP (M)
		&$\text{IS}_o$ (FPS)
		&$\text{IS}_d$ (FPS) \\ \hline
		BTS \cite{behind2023wimbauer}           &\textbf{43.92}	&\textbf{81.98}	&10.72 \\
		KYN \cite{know2024li}                   &530.49	&19.36	&1.03  \\ \hline
		\textbf{ViPOcc (Ours)}					&58.76	&81.68	&\textbf{124.58}\\
		\bottomrule
	\end{tabular}
	\caption{Comparison of network parameters and inference speed among ViPOcc and previous SoTA methods. TP denotes Training Parameters, $\text{IS}_o$ denotes the inference speed of 3D occupancy prediction, and $\text{IS}_d$ denotes the inference speed of depth estimation. }
	\label{tb.tp_is}
\end{table}

\subsection{Ray Sampler Efficiency}
Following previous studies \cite{know2024li, behind2023wimbauer}, we generate a total of 64 patches in each iteration, with each patch measuring $8\times8$ pixels. The previous random sampler often results in overlapped samples due to the absence of necessary constraints, leading to redundant computations for the same rays. This redundancy is considered unnecessary in our experiments. To evaluate the efficiency of our proposed SNOG sampler, we define $N_v$ as the average number of valid rays sampled per iteration, and $N_{vc}$ as the average number of valid rays sampled at crucial instances per iteration, which can be formulated as follows:

\begin{equation}
	N_v = \frac{1}{T}\sum_{j=1}^{j=T} \big|\{\boldsymbol{r}_{i,j}\}\big|, \ \ 
	N_{vc} = \frac{1}{T}\sum_{j=1}^{j=T} \big|\{\boldsymbol{r}_{i,j} | \boldsymbol{r}_{i,j}\in \mathcal{S}_j\}\big|,
\end{equation}
where $\boldsymbol{r}_{i,j}$ denotes the $i$-th sampled ray during the $j$-th iteration, $\mathcal{S}$ represents the set of rays associated with crucial instances, and $T$ denotes the total number of iterations, which is set to 1,000 in our experiments. 

Furthermore, we define $\psi_v$ as the average proportion of valid rays sampled at crucial instances versus the total number of valid rays, which is expressed as follows:
\begin{equation}
	\psi_v = \frac{1}{T}\sum_{j=1}^{j=T}\frac{\big|\{\boldsymbol{r}_{i,j} | \boldsymbol{r}_{i,j}\in \mathcal{S}_j\}\big|}{\big|\{\boldsymbol{r}_{i,j}\}\big|}\times 100\%.
\end{equation}
As illustrated in Table \ref{tb.snog_efficiency} and Fig. \ref{fig.results_sampler}, our proposed SNOG sampler not only achieves significant improvements in both $N_v$ and $\psi_v$ but also produces much more plausible sampled patches compared to the random sampler used by previous SoTA methods \cite{know2024li, behind2023wimbauer}. Our SNOG sampler can also be embedded into other SoTA methods to improve their 3D scene reconstruction performance. As shown in Table \ref{tb.snog_embedded}, both BTS \cite{behind2023wimbauer} and KYN \cite{know2024li} exhibit performance improvements when integrated with the SNOG sampler, primarily due to its instance-aware capabilities.

\begin{table}[t!]
	\centering
	\settablefont

	\begin{tabular}{c|cc}
		\toprule
		\multicolumn{1}{c|}{Ray Sampler} & $N_v$ ($\times10^3$) & $\psi_v$ (\%)\\
		\hline
		Random Sampler                  & 3.89  & 7.17  \\                          	
		SNOG Sampler (Ours)             & 4.10  & 36.83  \\
		\hline
		Improvement                     &5.40\% & 413.67\%    \\
		\bottomrule
	\end{tabular}
	\caption{Efficiency comparison between random sampler and our proposed SNOG sampler.}
	\label{tb.snog_efficiency}
\end{table}

\begin{table}[t!]
	\centering
	\settablefont

	\begin{tabular}{c|ccc|ccc}
		\toprule
		\multicolumn{1}{c|}{\multirow{2}*{Ray Sampler}}
		&\multicolumn{3}{c|}{BTS} &\multicolumn{3}{c}{KYN} \\
		\cline{2-7}
		& $\text{O}_{acc}^s$ 	
		&$\text{IE}_{acc}^s$
		&$\text{IE}_{rec}^s$
		& $\text{O}_{acc}^s$ 	
		&$\text{IE}_{acc}^s$
		&$\text{IE}_{rec}^s$ \\
		\hline
		w/o SNOG
		&0.91 &0.65 &0.64
		&0.92 &0.70 &0.66
		\\                          	
		w/ SNOG
		&0.92 &0.65 &0.65
		&0.92 &0.70 &0.67
		\\
		\bottomrule
	\end{tabular}
	\caption{Comparison between previous SoTA methods with and without our proposed SNOG sampler embedded.}
	\label{tb.snog_embedded}
\end{table}

\subsection{3D Occupancy Prediction}
We present additional qualitative experimental results of 3D occupancy prediction on the KITTI-360 dataset \cite{Kitti2022liao}. As illustrated in Fig. \ref{fig.results_occ_supp}, our method demonstrates the following advantages compared to previous SoTA methods:
1) superior object reconstruction performance in distant areas (shown in the first four rows), primarily due to the instance-aware sampling ray strategy in our proposed SNOG sampler;
2) fewer errors in 3D scene geometry recovery (shown in the last four rows), which can be attributed to the effective use of depth priors from the VFM and the enforcement of a reconstruction consistency constraint.

\subsection{Metric Depth Estimation}
We present additional qualitative experimental results of metric depth estimation on the KITTI-360 dataset \cite{Kitti2022liao}. As illustrated in Fig. \ref{fig.results_depth}, ViPOcc produces clearer boundaries of objects, such as trees, poles, and pedestrians, compared to previous SoTA methods BTS \cite{behind2023wimbauer} and KYN \cite{know2024li}. In addition, our method demonstrates superior depth consistency in continuous regions, evident in vehicle glass and road surfaces. These improvements are attributed to the efficacy of our proposed inverse depth alignment module, which not only effectively eliminates domain discrepancies but also preserves the fine-grained depth priors from the VFM.

\end{document}